\documentclass[11pt,a4paper]{article}
\usepackage[hyperref]{eacl2021}
\usepackage{times}
\usepackage{latexsym}

\usepackage{multirow}
\usepackage{textcomp}
\usepackage{graphicx} 
\usepackage{multicol} 
\usepackage{booktabs}
\usepackage{wrapfig}

\usepackage{microtype}

\aclfinalcopy 

\title{Domain-specific MT for Low-resource Languages: The case of Bambara - French}

\author{Allahsera Auguste Tapo\textsuperscript{1}, Michael Leventhal\textsuperscript{1}, Sarah Luger\textsuperscript{2} \\ \vspace{2mm}
{\bf Christopher M. Homan\textsuperscript{3}, Marcos Zampieri\textsuperscript{3}} \\

  \textsuperscript{1}Centre National Collaboratif de l’Education en Robotique et en Intelligence Artificielle (RobotsMali)\\
  \textsuperscript{2}Orange Silicon Valley,  \textsuperscript{3}Rochester Institute of Technology \\
  \texttt{aat3261@rit.edu} \\
  }

\date{}

\begin{document}
\maketitle
\begin{abstract}

Translating to and from low-resource languages is a challenge for machine translation (MT) systems due to a lack of parallel data. In this paper we address the issue of domain-specific MT for Bambara, an under-resourced Mande language spoken in Mali. We present the first domain-specific parallel dataset for MT of Bambara into and from French. We discuss challenges in working with small quantities of domain-specific data for a low-resource language and we present the results of machine learning experiments on this data.

\end{abstract}

\section{Introduction}

Most of the world's languages, of which approximately one-third are African, lack enough training data for conventional statistical and neural natural language processing (NLP) methods to apply.
An important development is the recognition that  NLP for such \emph{low-resource} languages requires its own practices, many of which can be shared among such languages. We have seen a number of recent initiatives in  this direction, such as Masakhane for African languages~\cite{masakhane} and annual Conference on Machine Translation (WMT) increasingly including low-resource languages in their competitions.
\cite{barrault2019findings,barrault2020findings}. 

In this paper we  present the a domain-specific parallel dataset for machine translation (MT) of Bambara into and from French. We discuss challenges in working with small quantities of domain-specific data for a low-resource language and we give insights as to how translation models trained with this data perform compared to the first models created for Bambara-French translation by \citet{tapo-etal-2020-neural}, by cross-testing models across domains. Our results will help to establish benchmarks for the performance of NMT models when bootstrapping an under-resourced language with small amounts of data from available sources as well as give insight into best practices with respect to data selection and cleaning.

\section{Background}

Bambara, a language from the Mande family in Western Sub-Saharan Africa, is spoken by approximately 16 million people. Thirty to forty million people speak a language from the Mande family. Bambara belongs to the dialect continuum of Manding languages in this family.~\cite{lewis2014ethnologue}.
It is the vehicular language of the nation of Mali and the first language of five million people and the second language of an additional eleven million. A majority of native speakers are from the Bambara ethnic group, spread throughout the African continent. There is also a large number of native-level speakers from many other ethnic groups coexisting in the region. Despite a substantial number of speakers, the language is extremely under-resourced, lacking a large quantity of parallel texts or even monolingual electronic texts. 

\citet{leventhal2020assessing, tapo-etal-2020-neural, luger2020towards} have been the first to research machine translation of Bambara. There has been a relatively large volume of linguistic research on the language itself~\citep{culy1985complexity,aplonova2017towards, aplonova2018development} which has included the preservation of virtually all published Bambara texts and a smaller number of parallel texts in Bambara and French in digital form.

\citet{tapo-etal-2020-neural} used a dictionary dataset from \emph{SIL Mali}\footnote{\url{https://www.sil-mali.org/en/content/introducing-sil-mali}} with examples of sentences in Bambara illustrating contextual word usage with full translations in French, English, and Spanish. This material can be generally characterised as intrinsically aligned since each sentence in Bambara was immediately followed by an exact translation. In addition, having the limited purpose of illustrating word usage, the sentences were independent of a larger narrative context which might have introduced elements in the text that could not be understood in the translation without that larger context. The material that we used in our experiments does not have these same qualities. While the dictionary contained only short sentences used in everyday Bambara speech, the domain-specific dataset consists of texts treating subjects that are only appear in specialised contexts related to health care, aligned at the chapter level and only impartially at paragraph and sentence levels. The objective of the translation in the dictionary data is to provide the exact meaning of the sentence in the destination language while the objective in the domain-specific dataset is to provide understanding of a medical topic that may not be possible to understand in the same way across linguistic and cultural differences.

\section{Data and Preprocessing}
\label{preprocessing}

\begin{figure}[!h]
    \centering
    \includegraphics[width=.48\textwidth]{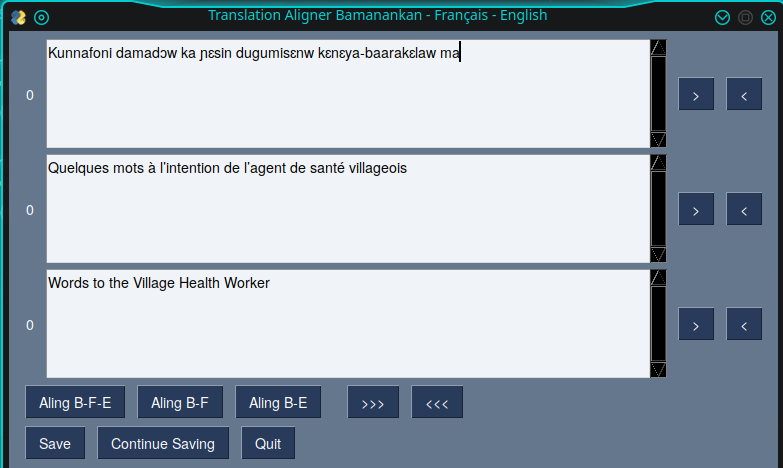}
    \caption[Custom aligner.]{The custom aligner reimplemented based on work previously done by \citet{tapo-etal-2020-neural} to manually align the medical dataset.
    }
    \label{fig:custom_aligner}
\end{figure}

\paragraph{Bambara Corpus}Our primary source was a tri-lingual (Bambara, French, English) rural health-care guide written for community health care workers entitled ``Where There Is No Doctor.\footnote{\url{https://gafe.dokotoro.org/}}'' The guide has been translated into many languages. The Bambara language version was created by The Dokotoro Project\footnote{\url{https://dokotoro.org/}} and used in this study with their permission. Detailed corpus statistics are listed in Table~\ref{tab:data_table}. We used a subset of the data, and only Bambara-French pairs. We plan to release the data publicly at \url{https://link_to_the_data.tdl}

\begin{table}[h]
\begin{center}
\scalebox{0.90}{
\begin{tabular}{cccc}
\toprule
& & \textbf{Bambara} & \textbf{French}  \\
\multirow{7}{*}{\rotatebox[origin=l]{90}{\textbf{Medical}}}  & chapters & \multicolumn{2}{c}{27} \\ 
 & files & \multicolumn{2}{c}{336} \\ 
 & paragraphs & 9,336 & 9,367 \\ 
  & unigrams & 8,209 & 9,893 \\ 
  & bigrams & 26,430 & 25,746 \\ 
 & trigrams & 5,816 & 11,312 \\ 
  & stopwords & 147 & 123 \\ 
  & \textbf{aligned} & 1,923 & 1,923 \\
 \bottomrule
\end{tabular}
}
\end{center}
\caption[Main data-sets.]{The dataset from the medical health guide with examples in Bambara, and French.}
\label{tab:data_table}
\end{table}


\paragraph{Sentence Alignment} We focus in this study only on passages of Bambara and their translations in French. While the authors of the medical health guide, as claimed in the introduction, aimed to write simply in order to make the text maximally accessible, the content nonetheless treats complex medical issues and is replete with medical terminology. In order to maintain narrative intelligibility, concepts described in one language are sometimes approached differently in the other. Consequently, only paragraph-to-paragraph alignment was dependable, with valid pairing in about 90\% of the cases. Sentence-to-sentence alignment within matched paragraphs worked in about 80\% of cases, which is to say that paired sentences corresponded in general meaning but were not necessarily considered ready for inclusion in the training set. Each sentence pair was evaluated by expert translators. A pair was rejected if there was significant information in one or the other sentence not found in its pair. As \citet{tapo-etal-2020-neural} previously discussed, the approximate nature of the alignments and differences in information content of paired sentences rendered fully automated pairing impossible. We found this to be true to an increased extent in our data, which compared to the data from \citet{tapo-etal-2020-neural} had longer sentences, more complex sentence compositions, and more domain-specific technical information.

Given this situation, we found that alignment could not be treated as a purely mechanical task nor one where expert knowledge of the two languages was sufficient. Automatic alignment was attempted using the MT-based sentence aligner Bleualign \citet{sennrich2010mt} but the results weren't sufficiently accurate to be useful in accelerating alignment. We did not attempt to engineer other possible automatic alignment approaches. Individual sentence pairs needed to be judged and, frequently, modified, taking into account linguistic and medical questions, and context. The availability of human resources able to perform this work was a major factor limiting the amount of data we were able to align during this phase of the project. To make the alignment work easier and more efficient to carry out, we use custom-built software shown in Figure \ref{fig:custom_aligner}. It enables annotators to manually align sentences or edit alignment pairings that another annotator already aligned or correct the content between aligned pairs to improve translation. In separate sessions and tasks, three annotators with formal education in written Bambara at the secondary school level performed alignment on Bambara-French sentence pairs, using the tool.

\paragraph{Preprocessing} Alignment was done in two steps, beginning with validation of alignments at the paragraph level from a subset of the medical text and other sources. Once paragraph alignments were produced, sentence alignments, paragraph by paragraph, were done manually. Data preparation, for which alignment was the great bulk of the work, proved to be about 90\% of the overall time spent in person-hours on the experiment. This aspect of the project required on-the-ground organisation and recruitment of skilled volunteers in Mali.

\paragraph{Parallel Data}
The final dataset contained 1,923 Bambara-French parallel sentences. While this represents fewer sentence pairs than the 2,146 pairs of \citep{tapo-etal-2020-neural}, the quantity of the text, shown in Table~\ref{tab:data_stats}, in number of words, is almost twice as large. 

\begin{table}[!ht]
    \centering
    \scalebox{0.78}{
    \begin{tabular}{llllll}
        \toprule
         &  &  & \textbf{Train} & \textbf{Dev} & \textbf{Test} \\
         \midrule
        \multirow{4}{*}{\rotatebox[origin=l]{90}{\textbf{Dict.}}} & \multirow{2}{*}{bm} & Avg. words/pair & 4.934 & 4.858 & 4.940 \\
         &  & Avg. uniq words/pair & 1.067 & 2.022 & 2.209 \\
         & \multirow{2}{*}{fr} & Avg. words/pair & \textbf{5.406} & \textbf{5.328} & \textbf{5.570} \\
         &  & Avg. uniq words/pair & \textbf{1.527} & \textbf{2.660} & \textbf{2.827} \\
         \midrule
        \multirow{4}{*}{\rotatebox[origin=l]{90}{\textbf{Med.}}} & \multirow{2}{*}{bm} & Avg. words/pair & 8.930 & 21.994 & \textbf{16.994} \\
         &  & Avg. uniq words/pair & 1.799 & 4.791 & 3.784 \\
         & \multirow{2}{*}{fr} & Avg. word/pair & \textbf{10.001} & \textbf{23.265} & 16.010 \\
         &  & Avg. uniq words/pair & \textbf{2.871} & \textbf{7.968} & \textbf{5.684} \\
         \midrule
        \multirow{4}{*}{\rotatebox[origin=l]{90}{\textbf{Combined}}} & \multirow{2}{*}{bm} & Avg. words/pair & 6.886 & 12.010 & \textbf{9.949} \\
         &  & Avg. uniq words/pair & 1.261 & 2.789 & 2.575 \\
         & \multirow{2}{*}{fr} & Avg. words/pair & \textbf{7.651} & \textbf{12.815} & 9.910 \\
         &  & Avg. uniq words/pair & \textbf{1.995} & \textbf{4.484} & \textbf{3.754} \\
        \bottomrule
    \end{tabular}
     }
    \caption{Data Characteristics (bold: highest value)}
    \label{tab:data_stats}
\end{table}

\begin{table*}[t]
    \centering
    \scalebox{0.90}{
    \begin{tabular}{lllllll}
        \toprule
        & & & \multicolumn{2}{c}{\textbf{bm$\rightarrow$fr}} & \multicolumn{2}{c}{\textbf{fr$\rightarrow$bm}} \\
         & \textbf{NMT Model} & \textbf{Configuration} & \textbf{BLEU} & \textbf{ChrF} & \textbf{BLEU} & \textbf{ChrF} \\
        \midrule
         (1) Dict. & BPE & BPE=1000 & \textbf{19.2} & 0.3 & \textbf{20.4} & 0.3 \\ 
         (2) Med. & BPE & BPE=1000 & 10.89 & 0.2 & 12.53 & 0.2 \\ 
         (3) Combined & BPE & BPE=1000 & 17.24 & 0.16 & 18.83 & 0.19 \\ 
         
         \midrule
         (4) Dict. & (1) + BPE dropout & dropout=0.1 & 16.0 & 0.3 & 17.7 & 0.2 \\ 
         (5) Med. & (2) + BPE dropout & dropout=0.1 & 8.03 & 0.2 & 6.16 & 0.2 \\ 
         (6) Combined & (3) + BPE dropout & dropout=0.1 & 19.15 & 0.15 & 19.01 & 0.16 \\ 
         
         \midrule
         \multicolumn{2}{l}{\textbf{Test scores} for the best models from above} &  & 20.9 & 0.3 & 21.2 & 0.2 \\ 
         \bottomrule
    \end{tabular}%
        }
    \caption{NMT results for French and Bambara translations in corpus BLEU and ChrF on the dev set. The best model (in bold) is evaluated on the test set, results are reported in the last row.}
    \label{tab:results_bm_fr}
\end{table*}

\begin{table*}[t]
    \centering
    \scalebox{0.80}{
    \begin{tabular}{cllllll}
        \toprule
        \multicolumn{1}{l}{} &  & \multicolumn{1}{c}{\textbf{}} & \multicolumn{2}{c}{\textbf{Dev}} & \multicolumn{2}{c}{\textbf{Test}} \\
        \multicolumn{1}{l}{} &  & \multicolumn{1}{c}{\textbf{}} & \multicolumn{1}{c}{\textbf{BLEU}} & \multicolumn{1}{c}{\textbf{ChrF}} & \multicolumn{1}{c}{\textbf{BLEU}} & \multicolumn{1}{c}{\textbf{ChrF}} \\
        \midrule
        \multirow{4}{*}{\textbf{Dict.'s models on Med. data}} & \multirow{2}{*}{\textbf{BPE=1000}} & {bm$\rightarrow$fr} & 1.53 & 0.08 & 3.82 & 0.10 \\
         &  & {fr$\rightarrow$bm} & 2.92 & 0.11 & 4.21 & 0.12 \\
         & \multirow{2}{*}{\textbf{BPE=1000 + dropout=0.1}} & {bm$\rightarrow$fr} & \textbf{4.79} & 0.09 & \textbf{8.7} & 0.10 \\
         &  & {fr$\rightarrow$bm} & 3.68 & 0.10 & 4.61 & 0.12 \\
         \midrule
        \multirow{4}{*}{\textbf{Med.’s models on Dict. data}} & \multirow{2}{*}{\textbf{BPE=1000}} & {bm$\rightarrow$fr} & 0.22 & 0.11 & 0.27 & 0.11 \\
         &  & {fr$\rightarrow$bm} & \textbf{0.32} & 0.10 & \textbf{0.94} & 0.10 \\
         & \multirow{2}{*}{\textbf{BPE=1000 + dropout=0.1}} & {bm$\rightarrow$fr} & 0.2 & 0.10 & 0.16 & 0.11 \\
         &  & {fr$\rightarrow$bm} & 0.23 & 0.08 & 0.56 & 0.09 \\
         \midrule
        \multirow{4}{*}{\textbf{Combined models on Med. data}} & \multirow{2}{*}{\textbf{BPE=1000}} & {bm$\rightarrow$fr} & 12.49 & 0.15 & 10.29 & 0.13 \\
         &  & {fr$\rightarrow$bm} & 14.23 & 0.19 & 8.52 & 0.13 \\
         & \multirow{2}{*}{\textbf{BPE=1000 + dropout=0.1}} & {bm$\rightarrow$fr} & 13.78 & 0.13 & \textbf{13.09} & 0.12 \\
         &  & {fr$\rightarrow$bm} & \textbf{14.32} & 0.16 & 10.10 & 0.12 \\
         \midrule
        \multirow{4}{*}{\textbf{Combined models on Dict. data}} & \multirow{2}{*}{\textbf{BPE=1000}} & {bm$\rightarrow$fr} & 8.65 & 0.19 & \textbf{10.12} & 0.19 \\
         &  & {fr$\rightarrow$bm} & \textbf{8.92} & 0.16 & 9.86 & 0.17 \\
         & \multirow{2}{*}{\textbf{BPE=1000 + dropout=0.1}} & {bm$\rightarrow$fr} & 7.53 & 0.17 & 7.51 & 0.16 \\
         &  & {fr$\rightarrow$bm} & 8.11 & 0.15 & 7.56 & 0.15 \\
         \bottomrule
    \end{tabular}
    }
    \caption{Cross Testing of Models and Datasets. (The best performing models' highest BLEU scores are in bold.)}
    \label{tab:cross_testings}
\end{table*}

\section{NMT Experiments}

We split the data randomly into training, validation, and test sets of 80\%, 10\% and 10\% respectively. The training set is composed of 1539 sentences, the validation set of 192 sentences, the test set of 190 sentences.
We utilized the same setup as \citet{tapo-etal-2020-neural}. Our NMT is a transformer~\citep{vaswani2017attention} of appropriate size for a rather smaller training dataset~\citep{biljon2020}. It contains six layers with four attention heads for the encoder and decoder, the transformer layer has a size of 1024, and the hidden layer size 256, the embeddings have 256 units. Embeddings and vocabularies are not shared across languages, but the softmax layer weights are tied to the output embedding weights.

The model is implemented with the Joey NMT framework~\citep{kreutzer-etal-2019-joey} based on PyTorch \citep{NIPS2019_9015}. Training was run for 120 epochs in batches of 1024 tokens each. The ADAM optimizer \citep{kingma2014adam} is used with a constant learning rate of 0.0004 to update model weights. This setting was found through tuning to yield the highest BLEU scores, as compared to decaying or warmup-cooldown learning rate scheduling. For regularization, we experimented with dropout and label smoothing. The best values were 0.1 for dropout and 0.2 for label smoothing across the board.
For inference, beam search with a width of 5 was used. The remaining hyperparameters are documented in the Joey NMT configuration files that is provided with the code.

Bambara lacks a standard tokenizer, like many other languages of the world \cite{tapo-etal-2020-neural}. The Bambara dev and test sets contain 920 and 720 distinct words, respectively. The French dev and test sets contain 1530 and 1080 distinct words, respectively. Because of this large proportion of unknown words, we segment the data for Bambara-French pair into subword units (byte pair encodings, BPE) (1000) using  subword-nmt\footnote{\url{https://github.com/rsennrich/subword-nmt}} \citep{sennrich-etal-2016-neural}, and apply BPE dropout to the training sets of both languages \citep{provilkov2019bpedropout}.

\section{Results}

\label{sec:auto-eval}
We evaluate the models' translations against reference translations on our heldout sets with corpus BLEU \citep{papineni2002bleu} and ChrF \citep{popovic2015chrf} computed with SacreBLEU~\citep{post-2018-call}.
Table~\ref{tab:results_bm_fr} shows the results for Bambara-French and French-Bambara translations respectively. We only focused on subword-level segmentation based on the experience of \citet{tapo-etal-2020-neural} which reported better results compared to word- and character-level models.

The BLEU and ChrF scores of the dictionary dataset previously studied are considerably higher than those of the medical dataset, as shown in Table~\ref{tab:results_bm_fr}. While the results on the two different datasets may not be directly comparable, the respective scores were not surprising given that the dictionary dataset consists mainly of simple and repetitive language patterns while the medical data is far more varied in the language constructs that are used. A model trained on data from both domains, however, performs about as well as the dictionary dataset alone. We were surprised by this result as we expected that the combining of a relatively limited dataset with a richer one would yield a net gain. Dropout yielded a relatively modest improvement, leaving open the question as to whether overfitting may or may not be a problem at this stage. The scores are generally low, either suggesting that the quantity of data remains too small or that BLEU and ChrF are not as useful for assessing results in this particular experiment or that the model is not tuned well enough, or some combination of these elements. Table~\ref{tab:cross_testings} confirms the natural intuition that running either the dictionary or medical dataset on the model of the other dataset produces very poor results and that running combined dataset on either model is less effective in terms of BLEU or ChrF scores than a model trained on combined data. It appears from our study that diversity is strength, that training on mixed styles of text yields better results in the early phases on developing an NMT system for an extremely low-resource language.

\section{Conclusion}
Our study constitutes the first attempt of modeling automatic translation for the extremely low-resource language of Bambara in a specific domain. We had more data available than we were able to use in our experiments due to the intrinsic challenges of alignment and the lack of alignment tools adapted to our language and the low-resource scenario. Future work may focus more on the alignment problem. The experiment has provided measures of content complexity and results as measured by BLEU and ChrF score which may be useful as a baseline for other projects working with small datasets for low-resource languages. The results illustrate a dimunation of BLEU and ChrF scores for more complex, domain-specific data. Results also suggest that combining domain-specific with more general data may improve overall performance on domain-specific data. Further experiments will add human evaluation to the assessment of transformer results. This may help to better interpret the significance of BLEU and ChrF scores on complex, domain-specific data.

\bibliography{anthology, references, eacl2021}
\bibliographystyle{acl_natbib}

\appendix

\end{document}